## Article

# The role of reciprocity in human-robot social influence

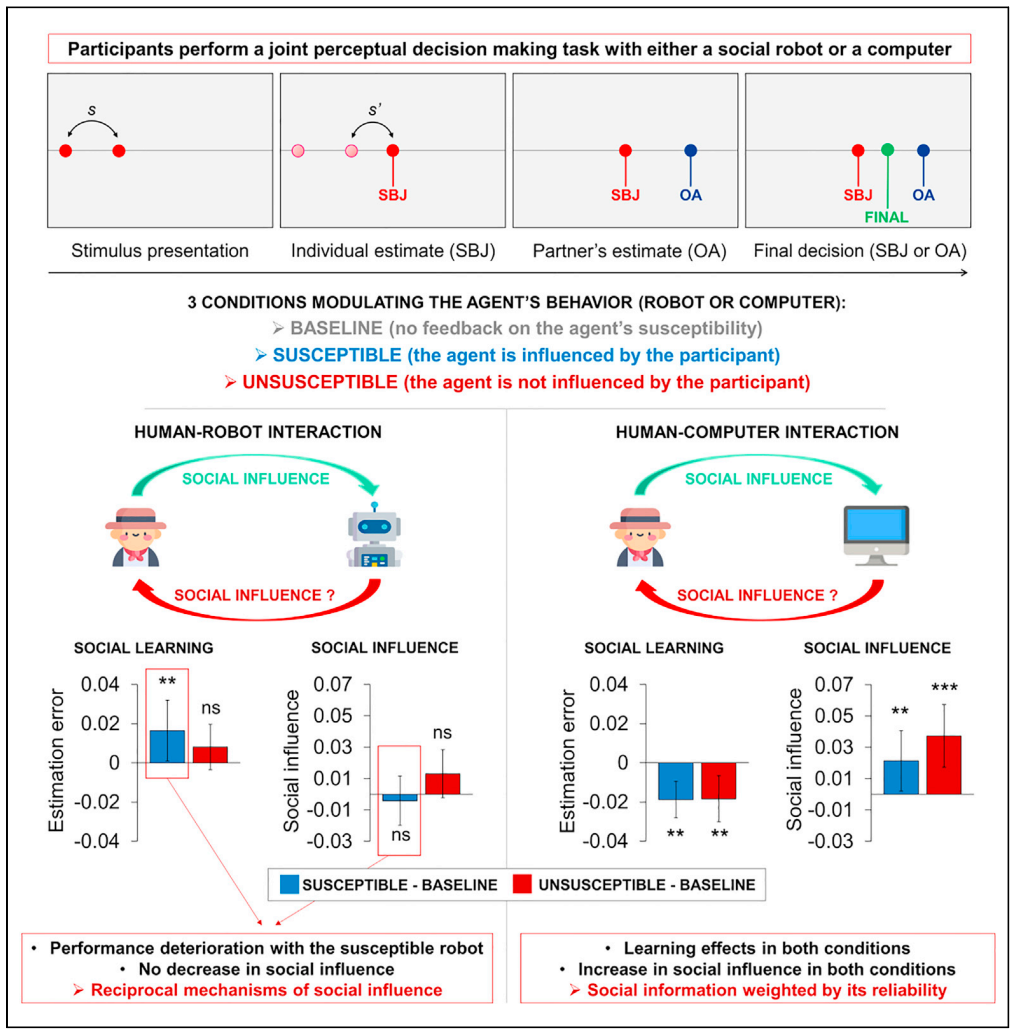

Joshua Zonca,
Anna Folsø,
Alessandra Sciutti

joshua.zonca@iit.it

**Highlights**

If a social robot is susceptible to our advice, we lose confidence in it

However, robot's susceptibility does not deteriorate social influence

These effects do not appear during interaction with a computer

Susceptible robots can promote reciprocity but also hinder social learning

## Article

# The role of reciprocity in human-robot social influence

Joshua Zonca,[1,3,*] Anna Folsø,[2] and Alessandra Sciutti[1]

## SUMMARY

**Humans are constantly influenced by others' behavior and opinions. Of importance, social influence among humans is shaped by reciprocity: we follow more the advice of someone who has been taking into consideration our opinions. In the current work, we investigate whether reciprocal social influence can emerge while interacting with a social humanoid robot. In a joint task, a human participant and a humanoid robot made perceptual estimates and then could overtly modify them after observing the partner's judgment. Results show that endowing the robot with the ability to express and modulate its own level of susceptibility to the human's judgments represented a double-edged sword. On the one hand, participants lost confidence in the robot's competence when the robot was following their advice; on the other hand, participants were unwilling to disclose their lack of confidence to the susceptible robot, suggesting the emergence of reciprocal mechanisms of social influence supporting human-robot collaboration.**

## INTRODUCTION

In the near future, robots will play a more and more important role in our society. In this regard, the dominant paradigm in the field of human-robot interaction (HRI) has been shifting from the ambition of building robots capable of autonomously accomplishing typical human tasks to the goal of designing robotic collaborators, which should be able to collaborate with humans in their everyday life. This ability would be decisive in a wide range of contexts, including education, industry, and care giving, in which the robots would assist human educators, workers, and operators in their activities. In order to achieve this goal, robots should be able to act as social entities on which human beings would rely on. Humans should be willing to accept help from robots and follow their advice in collaborative scenarios, as they do with their peers. This is not a trivial issue, since collaboration and cooperation, even among humans, are fragile processes that entail the presence and the persistence of different environmental and relational conditions. On the one hand, humans decide whether to follow others' behavior or advice based on their competence, which may be inferred by feedback on their reliability (De Martino et al., 2017; Park et al., 2017), confidence (Bahrami et al., 2010, 2012; Koriat, 2012; Sniezek and Van Swol, 2001), expertise (Bonaccio and Dalal, 2010; Boorman et al., 2013; Sniezek et al., 2004), and access to relevant information (Vélez and Gweon, 2019). On the other hand, individuals' reliance on peers follows inherently social and relational mechanisms entailing social norms, which lead humans to conform to their peers in order to affiliate with them and control reputation and self-image (Asch, 1951; Cialdini and Goldstein, 2014; Claidière and Whiten, 2012). In this regard, one of the most important social norms sustaining the evolution and the maintenance of cooperation is reciprocity (Nowak and Sigmund, 1998; Pfeiffer et al., 2005), which assumes that one's tendency to cooperate is conditional upon cooperative behavior of others (Barta et al., 2011; Gouldner, 1960; Nowak, 2006; Ohtsuki and Iwasa, 2004). Interestingly, recent evidence (Mahmoodi et al., 2018; Zonca et al., 2021a, 2021b) has shown that reciprocity also regulates social influence and compliance with others' judgments: people tend to take more into consideration the advice of peers who have previously taken into account their own opinion, even if, in principle, their susceptibility might signal incompetence or uncertainty.

Altogether, research in human-human interaction (HHI) highlights that, in social environments, individuals constantly balance informational and normative considerations in order to behave efficiently and, at the same time, preserve fruitful short- and long-term social relationships with their peers. Would this be the case also in interaction with (social) robots? Answering this question is still difficult, since research on social influence in HRI is rather limited. Research in HRI historically focused on the factors affecting humans' reliance on robots, which are conceptualized as mechanical agents performing actions or tasks, rather than intentional agents involved in symmetrical and reciprocal interactions. In this respect, numerous studies have shown that the main determinant of reliance on robots (often referred to as "trust," see Hancock et al., 2020) is its performance (Billings et al., 2012; Flook et al., 2019; Hancock et al., 2011; van den Brule et al., 2014; Wright et al., 2019). Humans tend to trust robots as long as they show reliable behavior, but they rapidly lose confidence in its competence in presence of failures (Desai et al., 2012; Rossi et al., 2017; Salomons et al., 2018), potentially leading to disuse of the robotic system (Lussier et al., 2007; Sanders et al., 2011). In the last years, research in HRI has been focusing on the introduction of a social dimension in human-machine interaction through the design of humanoid *social* robots, using HHI as behavioral model (Sandini and Sciutti, 2018; Strohkorb and Scassellati, 2016). The assumption is that robots appearing and behaving like humans might trigger the same types of emotional and behavioral reactions that are typically observed in HHI. In fact, recent evidence has shown the emergence of pro-social attitudes toward social robots (e.g., Connolly et al., 2020; Kahn et al., 2015; Kühnlenz et al., 2018; Siegel et al., 2009), which can enhance human-robot collaboration (e.g., Admoni and Scassellati, 2017; Baraglia et al., 2017; Oliveira et al., 2021; Terzioğlu et al., 2020). However, evidence of human-like relational mechanisms modulating HRI is rather limited and context dependent. Compliance toward robots tends to emerge primarily on functional tasks (i.e., tasks in which the robot has to fulfill a concrete goal through actions or quantitative judgment) rather than social ones (i.e., judgments or decisions on social issues), as shown in recent studies (Gaudiello et al., 2016; Ullman et al., 2021). Moreover, compliance is intrinsically tied with the uncertainty of the task solution: individuals typically conform to robots only if they are unsure about the action to take (Hertz and Wiese, 2018).

In sum, we have highlighted a fundamental difference between research in HRI and HHI: the former primarily focuses on humans' (functional) reliance on robots, which is determined by robots' capabilities; the latter constantly deals with normative and relational phenomenon (e.g., reciprocity) emerging between interacting peers. The main thesis of the current work is that, if we aim at building robots that could actively collaborate with humans in their everyday life, HRIs cannot be simply treated as a unidirectional relationship between a human agent and a task-oriented machine. Robots should identify us as living entities, try to anticipate our goals and needs, and engage with us on a social level. In other words, robots should act as social entities in bidirectional, reciprocal relationships with humans, especially in context in which robots should assist individuals in need (e.g., elderly people, individuals with reduced mobility). In the current study, we aim at enriching the study of social influence HRI by treating it as a relational and reciprocal process. We developed a novel experimental paradigm in which we are not solely interested in how the performance and the appearance of a robot can influence human reliance to robots. Indeed, we implemented a scenario in which the robot itself can express its own level of susceptibility (i.e., its tendency to change its opinion in favor of that of a human agent), paving the way for the (potential) emergence of reciprocal mechanisms of social influence, as already observed in recent studies in HHI (Mahmoodi et al., 2018; Zonca et al., 2021a, 2021b). Our paradigm consisted of three different tasks. In the Perceptual inference task, participants made individual perceptual judgments. In the Social influence task, participants made the same perceptual judgments and were told that either a humanoid robot iCub (Robot group) or a computer (Computer group) was doing the same. The participants received trial-by-trial feedback revealing the judgment performed by their partner. In each trial, after observing the agent's response, they could modify their original estimate by placing a final response between their own and their partner's estimate (final decision). In the Reciprocal social influence task, the task was the same but we alternated trials in which the final decision was taken by the participant (decision turns) with others in which the final decisions were made by the interacting agent (observation turns). The final decisions made by the agent in observation turns were manipulated to express different levels of susceptibility toward the participants (Susceptible condition: high susceptibility; Unsusceptible condition: low susceptibility). The partner's perceptual estimates and final decisions were controlled by the same algorithms in both the Robot and Computer groups. The participants were told that the agent would see both their perceptual estimates and final decisions. However, participants in the Robot group could not see their robotic partner during the tasks. In fact, we wanted the participants to acquire knowledge about the robot's behavior just by observing its (simulated) responses, in order to have full control of the robot's feedbacks and ensure the comparability with the Computer group. By controlling and systematically modulating the agent's level of susceptibility, we were able to analyze how the participants reacted and dynamically adjusted their own level of reliance on the partner's judgments. Nevertheless, in the Robot group, we wanted the participants to perceive the robot as a social and intentional agent, which was aware of the ongoing experiment with a human partner and could autonomously perform the joint task with them. To achieve this goal, participants in the Robot group, before starting the experimental session, had an encounter with the robot, which was programmed to perform specific verbal and motor actions (for a detailed description, see paragraph "introducing participants to the humanoid robot iCub" in the STAR Methods section).

The goal of this study is 2-fold. On the one hand, we explored how signals of susceptibility coming from a mechanical agent (robot or computer) have an impact on the willingness of the human partner to learn and take advice from it. For instance, we can hypothesize that a susceptible partner (computer or robot) signals uncertainty, which in turn may lead its partner(s) to rely less on its advice and help, as predicted by Bayesian theories of social information aggregation (Behrens et al., 2008; Boorman et al., 2013; De Martino et al., 2017). On the other hand, we investigated whether social influence during interaction with a social robot (as opposed to an *asocial* human-computer interaction) may be modulated by social and normative mechanisms. In particular, we tested whether the robot's susceptibility to the participant's judgements could be interpreted as a signal of a collaborative attitude during interaction, leading the participants themselves to reciprocate the consideration received from the robot in order to sustain cooperation with its robotic partner, following reciprocal processes typically intervening between human peers (Mahmoodi et al., 2018; Zonca et al., 2021a, 2021b).

## RESULTS

### Perceptual inference task

In each trial of the Perceptual inference task (Figure 1A), participants saw two consecutive red disks appearing for 200 ms on a visible horizontal line on a touch-screen tablet. The spatial distance between the two disks represented the target stimulus. Participants were then asked to touch a point to the right of the second disk to reproduce a segment matching the target stimulus length. No feedback about the accuracy of the response was provided. As a result of the well-known phenomenon of central tendency in quantity judgments (Hollingworth, 1910; Jazayeri and Shadlen 2010), participants' estimates should gravitate toward the mean magnitude of the visual stimuli (Figure 1B). Central tendency is an automatic perceptual mechanism aiming at error minimization in presence of sensory uncertainty, which can be represented by Bayesian models describing the mean magnitude of the stimulus history as a prior (Jazayeri and Shadlen 2010; Cicchini et al., 2012; Sciutti et al., 2014). Moreover, variability and uncertainty in perceptual estimation should increase along with stimulus magnitude, leading to a higher reliance on the Bayesian prior (i.e., mean magnitude of the stimulus history) in perceptual adjustment for longer stimuli (Petzschner et al., 2015). These mechanisms are expected to lead to a general underestimation of lengths (Figure 1B).

First, we quantified the effect of central tendency by computing the *regression index*, defined as the difference in slope between the identity line (i.e., veridical reproduction of stimuli) and the best linear fit of the reproduced lengths. Therefore, a regression index approaching 1 reflects complete regression to the mean, whereas a regression index approaching 0 expresses no effect of central tendency. To explore the effect of central tendency within and across groups, we ran a linear regression with regression index as dependent variable and group (Computer or Robot) as independent factor. Results reveal that the regression index is significantly different from 0 in both groups (intercept, Computer group: B = 0.46, t = 9.79, p < 0.001; Robot group: B = 0.38, t = 8.04, p < 0.001), with no significant interaction (Robot − Computer: B = −0.08, t = −1.23, p = 0.223). Moreover, we tested the hypothesis that participants underestimated lengths by running a linear regression with the difference between the mean of the reproduced stimuli and the mean of actual stimulus distribution (= 12 cm) as dependent variable, and group and independent factor. The results reveal a significant effect of underestimation in both groups (intercept, Computer group: B = −1.52, t = −5.75, p < 0.001; Robot group: B = −0.92, t = −3.47, p < 0.001), with no significant interaction (Robot − Computer: B = 0.60, t = 1.61, p = 0.113). Altogether, these results indicate the presence of significant effects of central tendency and underestimation of the reproduced lengths of similar magnitude in both experimental groups (Figure 1C).

Then we further tested the comparability of our two experimental groups by investigating the presence of a between-group difference in terms of trial-by-trial *estimation error*, computed as the absolute difference (in centimeters) between the estimated length and the actual stimulus length divided by the actual stimulus length. This parameter is directly related to central tendency and underestimation of lengths and is particularly important because it directly expresses participants' accuracy, which will have a fundamental role in the analysis of the subsequent interactive tasks. Indeed, a between-group difference in perceptual accuracy in the Perceptual inference task could represent a confound factor for the analysis of social influence and perceptual learning in human-robot and human-computer interactions. Nonetheless, results show that

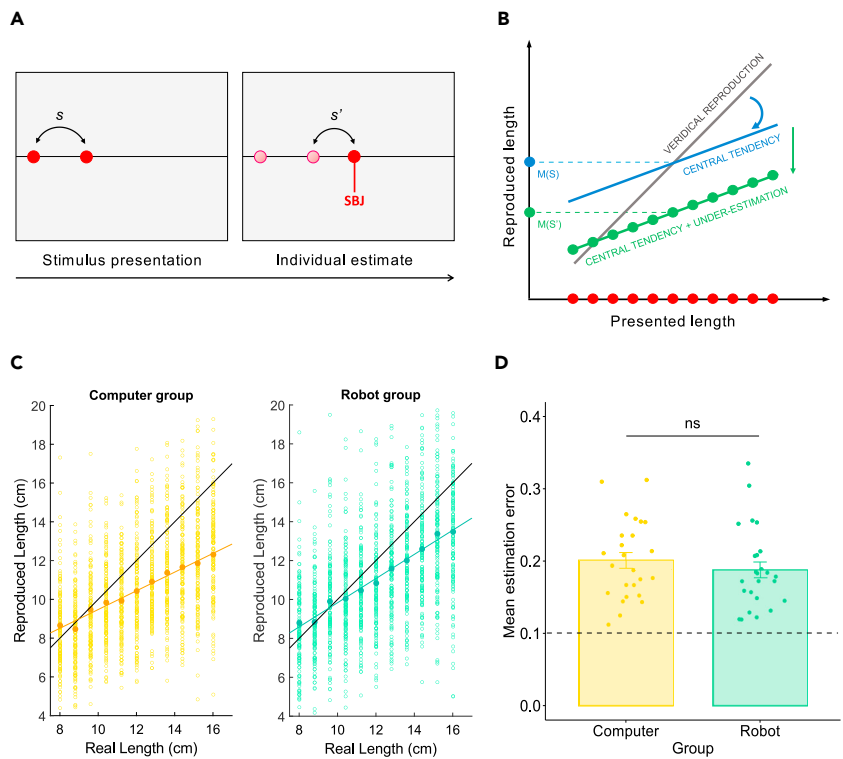

**Figure 1. Perceptual inference task**

(A) Perceptual inference task. Participants saw two red disks appearing consecutively with a duration of 200 ms. Participants had to touch a point, to the right of the second disk, in order to reproduce the stimulus length (s'), defined as the distance between the first and the second disk (s).

(B) Expected perceptual phenomena in the Perceptual inference task. Owing to central tendency, participants could be expected to underestimate long stimuli and overestimate short stimuli. Moreover, we expect the central tendency effect to be stronger for long distances, leading to a general underestimation of the mean of the reproduced length distribution (M(S')).

(C) Responses in the Perceptual inference task in the Computer and Robot groups. Gold (Computer group) and light green (Robot group) dots represent participants' reproduced lengths for each presented length. Orange (Computer group) and dark green (Robot group) dots represent the participants' mean reproduced length for each presented stimulus. Orange and dark green lines represent the linear fit between the presented and reproduced lengths in the Computer and Robot groups, respectively. Black lines represent the identity line (veridical reproduction). The pattern of responses in both groups reflects the emergence of the perceptual mechanisms depicted in Figure 1B. In both groups, the slope of the reproduced lengths is flatter than the identity line, indicating central tendency. Moreover, we observe that the central tendency is markedly more pronounced for long stimuli, resulting in a general underestimation of the presented visual stimuli.

(D) Average estimation error in the Computer and Robot groups. The two experimental groups are comparable in terms of estimation error, which is computed as distance (in cm) from the correct response/stimulus length (ns: not significant, Wilcoxon rank-sum test). The dotted line represents the average estimation error of the computer algorithm acting as the participants' partner in both the Computer and Robot groups in the Social influence task (see next paragraph). All participants show a higher mean estimation error than the computer algorithm. Error bars represent between-subject standard error of the mean, and gold/green dots indicate individuals' mean estimation error. ns: not significant, Wilcoxon rank-sum test.

the two groups did not differ in terms of average estimation accuracy (mean estimation error, Computer group: $0.20 \pm 0.05$; Robot group: $0.19 \pm 0.06$. Robot – Computer: Wilcoxon rank-sum test, $z = 1.00$, $r = 0.14$, $\eta^2 = 0.02$, $p = 0.318$. Figure 1D).

## Social influence task

In the Social influence task (Figure 2), in each trial, participants made their perceptual estimate as in the Perceptual inference task and then observed the estimate (concerning the same stimulus) made by a partner (robot in the Robot group, computer in the Computer group). This feedback consisted of a vertical line

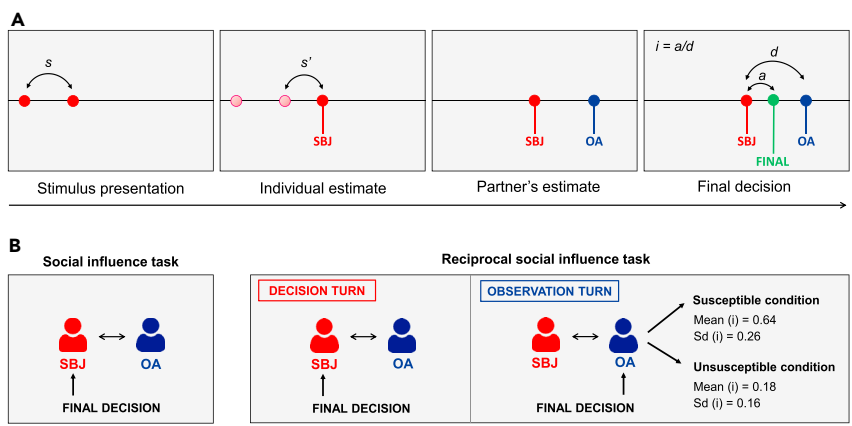

**Figure 2. Social influence task and Reciprocal social influence task**

(A) Experimental task. In both the Social influence task and Reciprocal social influence task, initially participants had to reproduce the lengths of visual stimuli, as in the Perceptual inference task. Participants were told that the very same stimuli would have been presented also to their partner (other agent: OA), which was a humanoid robot in the Robot group and a computer in the Computer group. Participants were told that the partner would choose a point to reproduce the length of the presented stimulus. After the participants' estimate, the partner's simulated estimate was shown and then one of the two agents had the opportunity to make a final decision by choosing any position between own and partner's estimates. The index of influence (i) was defined as the adjustment toward the partner (a) divided by the distance between the two agents' responses (d).

(B) Experimental design. In the Social influence task, participants made final decisions in all trials. In the Reciprocal social influence task, in half of the trials (decision turns) participants performed the same task with the same partner and made final decisions. Decision turns were alternated with observation turns, in which the partner made the final decision, while participants could not revise their estimate. We manipulated the partner's final decisions across two experimental conditions to express two different levels of susceptibility toward participants' responses (Susceptible: high influence; Unsusceptible: low influence).

indicating the exact location of the partner's estimate on the tablet. Afterward, participants were asked to make a final decision by choosing any position between their own and the partner's response. Participants were incentivized to be as accurate as possible in both perceptual estimates and final decisions (see the paragraph "overview: participants and procedure" in the STAR Methods). Participants were also told that their partner could see their final decisions. The shift from their own estimate to that of the partner (divided by the distance between the two estimates) has been used as an index of *influence* from the partner, which expressed the relative weight assigned by participants to either their own or the partner's estimate. At the end of the task, participants evaluated (1–10) their own and the other agent's accuracy in perceptual estimation to obtain a subjective measure of perceived competence.

First, it is important to highlight that participants' accuracy in terms of perceptual estimates was remarkably lower than that of their partner in both experimental groups (Wilcoxon rank-sum test on estimation error. Computer group: $z = 6.06$, $r = 0.86$, $\eta^2 = 0.73$, $p < 0.001$; Robot group: $z = 6.06$, $r = 0.86$, $\eta^2 = 0.73$, $p < 0.001$. Results significant at the Bonferroni-corrected threshold for two comparisons). Participants' estimation error was comparable across experimental groups (Wilcoxon rank-sum test, $z = 1.04$, $r = 0.15$, $\eta^2 = 0.02$, $p = 0.299$). Moreover, participants' estimation error was not different from that observed in the Perceptual inference task in both groups (Wilcoxon signed-rank test on estimation error. Computer group: $z = 0.040$, $r = 0.01$, $\eta^2 = 0.00$, $p = 0.968$; *Robot* group: $z = -0.740$, $r = 0.10$, $\eta^2 = 0.01$, $p = 0.459$. Figure 3A), suggesting that the partner's perceptual estimate was not used as a feedback for learning. The observed inability to use the partner as a valuable information source for learning is indeed linked to a marked distortion in the participants' perception of their own and others' performance. Specifically, participants were not able to recognize the higher accuracy of the partner, since their performance ratings were even higher for own accuracy (Computer group: $6.28 \pm 1.10$; Robot: $6.4 \pm 1.08$) than their partner's one (Computer group: $5.52 \pm 1.64$; Robot group: $6.24 \pm 1.90$). See Figure 3C).

In line with the observed overestimation of own performance and underestimation of the partner's one, analysis of participants' final decisions reveals that participants took into account the partner's response much less than their own estimate (Average influence, Computer group: $0.26 \pm 0.13$; Robot group: $0.35 \pm 0.16$. Wilcoxon

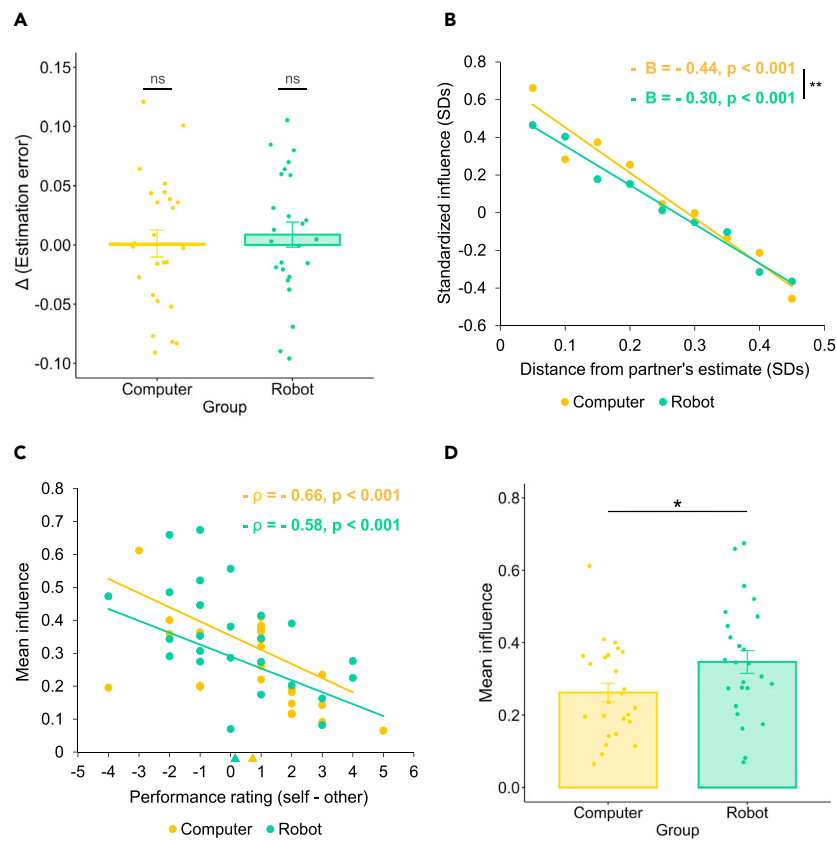

(A) Estimation error across Perceptual inference task and Social influence task. We plot the difference between estimation error in the Social influence task and the Perceptual inference task, highlighting the absence of learning effects between the two tasks in both the Robot and Computer groups. Error bars represent between-subject standard error of the mean, and gold/green dots indicate individuals' mean estimation error. ns: not significant, Wilcoxon signed-rank test.

(B) Influence as a function of agents' response distance. For the purpose of this graph, first we calculated the standard deviation of participants' estimates for each stimulus length. Then, for each stimulus, we grouped trials based on the distance of the partner's estimate from the partner's one using the standard deviation of the participant's estimates of that stimulus as unit of measurement. We used nine distance ranges with a step of 0.05 SDs. For instance, we grouped in the first distance range all trials in which the distance from the partner' estimate was less than 0.05 SDs (considering 1 SD as the standard deviation of the participants' estimates of the current stimulus). The next range included distances ranging from 0.05 to 0.1 SDs, and so on. Then, for each distance range, we calculated the average standardized participant's influence and finally we averaged these indices across participants. Influence was standardized to express within-subject variation of influence as a function of the distance from the partner's response, taking into account individuals' level of influence. Gold and green lines represent linear fits in the two groups. The statistics (B and p) reported in the figure refer to the (unstandardized) coefficients and the p values of model 1 (supplemental information, model 1, Table S1), which investigates the relationship between influence and agents' response distance across groups. **p < 0.05, group*distance interaction effect on influence in model 1.

(C) Influence and subjective performance assessment. Mean influence plotted as a function of individual performance rating (self – other) in the Social influence task. Gold and green lines represent linear fits in Computer and Robot conditions, respectively. Gold and green triangles on the x axis indicate mean ratings for the Computer and Robot groups, respectively. The scatterplot reports results of Spearman correlations (ρ and p value) in both groups.

(D) Influence in robot and computer partners. Bar plot of mean influence in the Computer and Robot groups. Error bars represent between-subject standard error of the mean. Participants in both groups relied more on their own estimate than on that of the partner (influence <0.5, p < 0.001, Wilcoxon signed-rank test). Influence was higher in the Robot than in the Computer group. *p < 0.05, Wilcoxon rank-sum test.

signed rank-test, null hypothesis: average influence = 0.5. Computer group: z = 4.29, r = 0.86, $\eta^2$ = 0.74, p < 0.001; Robot group: z = 3.51, r = 0.70, $\eta^2$ = 0.49, p < 0.001. Results significant at the Bonferroni-corrected threshold for two comparisons). Participants' average influence was significantly correlated with performance ratings (i.e., self

– other rating. Spearman correlation, Computer group: rho = −0.66, p < 0.001; Robot group: −0.58, p = 0.002. Results significant at the Bonferroni-corrected threshold for two comparisons; Figure 3C), revealing that final decisions strongly depended on the perceived competence of the partner. Interestingly, we observed that participants in the Robot group were more influenced by their partner than participants in the Computer group (Wilcoxon rank-sum test of average influence, z = −1.99, r = 0.28, $\eta^2$ = 0.08, p = 0.047; Figure 3D). Altogether, these findings suggest that participants in both groups, in the absence of direct or indirect feedback revealing the accuracy of the two agents, overestimated their own ability and underestimated that of the partner, which prevented participants from learning and led to low reliance on its judgments. This effect was more pronounced when participants believed the partner to be a computer rather than a robot.

Then we investigated whether within-subject modulation of social influence was linked to the perceived reliability of the partner's feedback. We tested the relationship between influence and distance from the partner's response, which was the only feedback available to participants for inferring own and other's performance. We ran a mixed-effect linear regression with influence as dependent variable and group, distance, and their interactions as independent variables, with subject as random effect (supplemental information, model 1, Table S1). Results show an effect of distance on influence in both Computer (B = −0.44, z = −12.33, p < 0.001) and Robot (B = −0.30, z = −9.36, p < 0.001) groups: more specifically, participants' influence significantly decreased with the increase of the distance between the two estimates (Figure 3B). On the one hand, this effect suggests that participants' influence was modulated by the perceived reliability of the current partner's feedback; on the other hand, it reveals that the weight assigned to the partner's judgment was weighted less and less as the discrepancy between the two interacting partners increased. We highlight that the effect is significantly stronger when participants believed they were interacting with a computer (interaction effect, B = 0.14, z = 2.87, p = 0.004), suggesting that susceptibility to the computer partner decreased faster (as a function of agents' response distance) than that to the robot partner.

## Reciprocal social influence task

The Reciprocal social influence task (Figure 2) consisted of two different types of experimental trial: in *decision* turns, the task was identical to the Social influence task and the participant took the final decision after observing the partner's estimate. As in the Social influence task, participants were told that their partner could observe both their perceptual estimates and final decisions and were incentivized to be as accurate as possible in both perceptual estimates and final decisions. In *observation* turns, the partner itself (robot or computer) made the final decision and the participant could just observe the partner's choice. Decision and observation turns were continuously alternated along the task. In observation turns, the susceptibility of the agent toward the participant's estimates (i.e., the influence index in its final decisions) was manipulated in two different conditions (Susceptible: high influence; Unsusceptible: low influence). See paragraph "agents' final decisions in the reciprocal social influence task" in the STAR Methods for a detailed description of the relevant algorithms. We aimed at analyzing if participants' influence was affected by the influence shown by the partner. In this respect, we highlight that participants' performance ratings (self – other) in the Reciprocal social influence task conditions depended on the *partner's* level of influence in the current condition (B = 1.75, z = 3.70, p < 0.001; supplemental information, model 2, Table S2). This result reveals that providing feedback about the partner's susceptibility did have an impact on the perceived reliability of the partner.

First, we tested whether the partner's feedback promoted or disrupted participants' learning (Figure 4). We compared participants' estimation error in Susceptible and Unsusceptible conditions with that observed in the Social influence task, which will serve as a "Baseline" condition. We ran a mixed-effects model with trial-by-trial estimation error as dependent variable, experimental group (Computer or Robot) and experimental condition (Susceptible, Unsusceptible and Baseline) and their interactions as predictors and a random effect on the intercept at the subject level (supplemental information, model 3, Table S3). Results reveal a significant learning effect (i.e., decrease in estimation error) in the Computer group in both conditions (Susceptible – Baseline: B = −0.02, z = −2.75, p = 0.006; Unsusceptible – Baseline: B = −0.02, z = −2.67, p = 0.008). Interaction effects reveal that the effect of learning in the Computer group was more pronounced than in the Robot group (Susceptible – Baseline, Robot – Computer: B = 0.03, z = 4.00, p < 0.001; Unsusceptible – Baseline, Robot – Computer: B = 0.02, z = 2.90, p = 0.004). In fact, we do not observe an effect of learning for the Robot group in the Unsusceptible condition (Unsusceptible – Baseline: B = 0.01, z = 1.43, p = 0.154) and we even observe a significant *increase* in estimation error in the

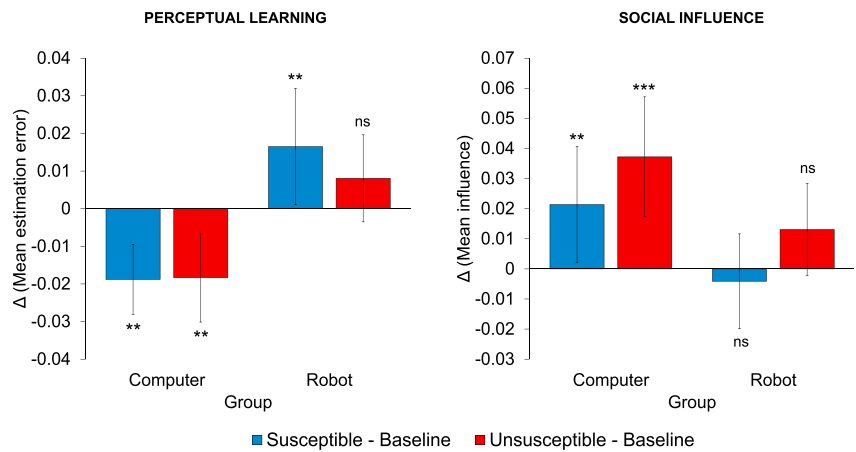

**Figure 4. Results of the Reciprocal social influence task**

Bar plots of mean estimation error (left panel) and mean influence (right panel) depending on experimental group (Computer or Robot) and experimental condition (Susceptible or Unsusceptible). All measures were compared with the "baseline" level of estimation error and influence expressed in the Social influence task. For example, the first blue bar in the left panel refers to the mean estimation error in the Susceptible condition of the Reciprocal social influence task minus the mean estimation error in the Social influence task, in the Computer group. Error bars represent between-subject standard error of the mean. \*\*\*p < 0.001, \*\*p < 0.01, ns: not significant, mixed-effects model (model 4 for the "perceptual learning" analysis, model 5 for the "Social influence" analysis. See supplemental information for the model specifications and their complete results in Tables S4 and S5).

Susceptible condition in the Robot group (Susceptible – Baseline: B = 0.02, z = 2.91, p = 0.004). The latter effect was explained by a significant relationship between deterioration of performance in the Susceptible condition and increase in the discrepancy between own and agent's performance ratings in the same condition, which was present in the Robot group but not in the Computer group (linear regression, effect of ratings in the Robot group: B = −0.04, t = −4.01, p < 0.001; effect of ratings in the Computer group: B = 0.00, t = 0.27, p = 0.785; interaction effect, Computer − Robot: B = 0.04, t = 3.00, p = 0.004).

We also analyzed changes in terms of distance from the partner's response to understand whether the learning patterns observed in the two groups were related to a systematic shift in the response distribution toward (or away from) that of the partner across tasks and conditions. Therefore, we ran the same model of the previous analysis using trial-by-trial response distance (normalized by the current stimulus length) as dependent variable (supplemental information, model 4, Table S4). Results confirm that the learning effects in the Computer group were accompanied by a decrease in the distance from the partner's response (Susceptible – Baseline: B = −0.01, z = −2.30, p = 0.021; Unsusceptible – Baseline: B = −0.02, z = −3.42, p = 0.001). Interaction effects reveal that the effects observed in the Computer group were more pronounced than in the Robot group (Susceptible – Baseline, Robot – Computer: B = 0.03, z = 3.61, p < 0.001; Unsusceptible – Baseline, Robot – Computer: B = 0.03, z = 3.35, p = 0.001). Indeed, we do not see an effect of response distance for the Robot group in the Unsusceptible condition (Unsusceptible – Baseline: B = 0.01, z = 1.32, p = 0.188) and we observe a significant *increase* in response distance in the Susceptible condition in the Robot group (Susceptible – Baseline: B = 0.02, z = 2.80, p = 0.005). Taken together, these results suggest that participants in the Computer group did use the partner's susceptibility feedback as a learning signal that helped them in improving their perceptual performance, whereas participants in the Robot group interpreted their partner's susceptibility as evidence of its incompetence, which disrupted social learning and led to a deterioration of behavioral performance when the susceptibility expressed by the partner was high.

Then we analyzed participants' final decisions to test whether the presence of feedback about the partner's susceptibility shaped the participants' expressed susceptibility to their partner's judgments (Figure 4). A first hypothesis is that results of participants' final decisions, which express an *overt* signal of susceptibility to the human partner, mirror the learning effects observed in Reciprocal social influence task, which in turn represent a *covert* signal of participants' reliance on the partner's judgments linked to its perceived competence. More specifically, participants in the Computer group, who used the feedback on the partner's susceptibility as a feedback supporting learning, should *increase* their level of influence in their final decisions. Conversely, participants in the *Robot* condition should decrease their influence level in the Susceptible condition, whereas they should not change their level of influence in the Unsusceptible condition. We ran a mixed-effects model with trial-by-trial influence in decision trials as dependent variable, experimental group (Computer or Robot), and experimental condition (Susceptible, Unsusceptible and Baseline) and their interactions as predictors and a random effect on the intercept at the subject level (supplemental information, model 5, Table S5). In the Computer group, results indeed reveal a significant increase of influence in both conditions (Susceptible – Baseline: B = 0.03, z = 2.95, p = 0.003; Unsusceptible – Baseline: B = 0.04, z = 4.74, p < 0.001), in line with the previously observed learning effects. We also observe a significant interaction with the Robot group in both conditions (Susceptible – Baseline, Robot – Computer: B = −0.03, z = −2.41, p = 0.016; Unsusceptible – Baseline, Robot – Computer: B = −0.03, z = −2.31, p = 0.021), revealing that the effect of influence was more pronounced in the Computer group. In fact, we do not observe any effect of condition in the Robot group (Susceptible – Baseline: B = −0.00, z = −0.46, p = 0.644; Unsusceptible – Baseline: B = 0.01, z = 1.48, p = 0.140). This latter result reveals an inconsistency between the observed covert decrease in susceptibility toward the susceptible robot partner, as signaled by the detrimental effect in terms of participants' perceptual learning, and the lack of an equivalent decrease in the overt susceptibility to the robot expressed through the final decisions. We may interpret this inconsistency by assuming that the susceptible robot, showing a high level of susceptibility to the participant, conveys a socially positive signal that is taken into consideration by the participants themselves in their final decisions. In particular, the participant, despite the loss of confidence in the robot competence, is not willing to anti-reciprocate the consideration received from the robot by explicitly decreasing the magnitude of the perceptual adjustments in their final decisions.

In order to better characterize these effects of reciprocity in the Robot group, we also looked at the direct differences between Susceptible and Unsusceptible conditions in model 5. In this regard, previous results in dyadic HHIs (Mahmoodi et al., 2018; Zonca et al., 2021b) have revealed that humans take *more* into account the judgment of a peer if the latter has taken into account their opinion (i.e., participants show higher influence in the Susceptible than in the Unsusceptible condition), in line with a normative effect of reciprocity. Nonetheless, results of model 5 show that the level of reciprocity observed in HHI does not emerge in either the Computer group (Susceptible – Unsusceptible: B = −0.02, z = −1.55, p = 0.121) or in the Robot group (Susceptible – Unsusceptible: B = −0.02, z = −1.68, p = 0.093), with no interaction between the two groups (Susceptible – Unsusceptible, Robot – Computer: B = −0.00, z = −0.09, p = 0.927).

In this regard, the effect of reciprocity observed in human dyadic and triadic relationships (Mahmoodi et al., 2018; Zonca et al., 2021a) has been linked to the desire of maintaining influence over the interacting partner, given the positive reward signal provided by a high level of consideration received from them. Therefore, we tested whether this phenomenon could explain the effects of reciprocity observed in the Robot group. We explored this hypothesis by introducing, at the end of the Reciprocal social influence task, an additional experimental block (i.e., Final block) in which participants were explicitly told that they would perform 11 observation turns in a row followed by 11 decision trials in a row. Importantly, participants were informed that the experiment would finish after the 11 decision trials. If, during the Reciprocal social influence task, participants' level of influence was indeed modulated by a mechanism of reciprocation directed to maintain influence over the partner, we should observe a decrease in participants' influence in the Final block, where there is no expectation of future interactions with the partner. To explore this hypothesis, we compared influence in the Final block with influence in the preceding main experimental block: these two blocks indeed coincide in terms of behavior of the agent but differ in the prospect of future interactions with the active partner. Results of a mixed-effect model (supplemental information, model 6, Table S6) show a significant drop in participants' influence in the Final block in the Robot group (B = − 0.034, z = − 2.45, p = 0.014), which was absent in the Computer group (B = − 0.016, z = − 1.14, p = 0.255). However, the interaction term was not significant (B = 0.018, z = 0.93, p = 0.354), which prevents us from drawing conclusions on the role of the nature of the partner on participants' behavior in the Final block.

## DISCUSSION

Recent studies on human social behavior have shown that social influence among peers is modulated by reciprocity: we tend to rely more on others' opinions if others take into consideration our ideas (Mahmoodi et al., 2018; Zonca et al., 2021a, 2021b; Zonca et al., 2021a). Conversely, research in HRI has historically conceptualized human-robot relationships as unidirectional, treating them as human-machine interactions in which humans' reliance on robots depends (almost) exclusively on their performance. We are persuaded that reciprocity may play a role in the emergence of (mutual) social influence also in human-robot collaboration. Therefore, we designed a novel experimental paradigm, composed of three different tasks, meant to investigate the emergence of mechanisms of reciprocal social influence in HRI. In the Perceptual inference task participants had to make perceptual judgments individually. In the Social influence task, participants made the same perceptual inferences but, in each trial, they could also observe the estimate of another agent (a humanoid social robot in the Robot group, a computer in the Computer group). The estimates of the two agents were systematically controlled and identical across groups. In each trial, participants were then asked to make a final decision by selecting a position between the two responses, trying to maximize accuracy. The shift from their own original estimate to that of the partner has been used as an index of *influence* from the partner. Results reveal that, in both groups, participants did not realize that their own accuracy was markedly lower than their partner's one: the average level of influence was generally low (i.e., <0.5) and rapidly decreased as the distance from the other's estimate increased, highlighting the presence of egocentric distortions in competence assessment, as commonly observed in HHI (e.g., Morin et al., 2021; Yaniv, 2004; Yaniv and Kleinberger, 2000). Moreover, their performance did not improve with respect to the Perceptual inference task, suggesting that the partner's feedback was not used as a valuable signal for perceptual learning. These findings shed new light on humans' disposition toward the advices of mechanical agents in the context of perceptual estimation. In the absence of feedback about own and other's performance, participants were not able to attribute a correct weight to the partner's advices, suggesting that humans may necessarily need empirical evidence about robots' and computers' performance to rely on their (perceptual) judgments in a joint task, as in the case of social influence among peers. Interestingly, participants' susceptibility to the partner's judgment was significantly higher in the Robot group than in the Computer group, although their task-related competences were identical. This result underlines the importance of prior beliefs and dispositions about the nature and the capabilities of mechanical agents in determining reliance on them (Kaniarasu et al., 2013; Xu and Dudek, 2016).

Then we investigated whether and how the susceptibility to the computer or robot partner varied in context of reciprocal social influence (Reciprocal social influence task). Participants in both the Computer and Robot groups alternated trials in which they personally had to make the final decision (decision turns) with trials in which the final decision was made by the partner (observation turns). We systematically manipulated the level of influence of the partner in two different conditions (Susceptible: high susceptibility; Unsusceptible: low susceptibility) and analyzed participants' behavior in decision turns by comparing choices in Susceptible and Unsusceptible conditions with behavior shown in the Social influence task.

First, results of the Computer group reveal a significant effect of learning in both conditions of the Reciprocal social influence task compared with the Social influence task. This suggests that receiving information about the current level of susceptibility of the computerized agent led participants to increase their general level of confidence in the accuracy of the computer, eliciting learning processes. In line with this interpretation, participants in the Computer group did also increase the level of influence expressed in their final decisions in both conditions of the Reciprocal social influence task. We hypothesize that participants may have started to rely more on the feedback of the computer algorithm owing to its tendency to variably balance own and others' perceptual judgments based on an internal weighting model. The exposure to this decision strategy may have led the participants to reinforce the idea that a balance between own and partner's divergent estimation patterns could improve the final performance. Nonetheless, our results highlight that these effects did not overturn the egocentric discounting observed in the Social influence task: participants still relied more on their own perceptual judgments than those of the partner, even if the latter was more accurate than them.

Contrary to the learning effects observed in the Computer group, participants in the Robot group *did not* benefit from the feedback about the robot's susceptibility in terms of perceptual learning. Their performance even deteriorated in the Susceptible condition, when the robot was relying heavily on participants' estimates in its final decisions. This result indicates that the susceptible robot conveyed a signal of uncertainty or incompetence, leading to a re-calibration of the human's perceived reliability of the robot in the presence of response discrepancy. The subsequent loss of confidence in the robot's competence is consistent with the idea that humans, although possessing relatively high prior confidence in robotic technology, easily lose confidence in its contingent capacity if they receive evidence signaling unreliability or uncertainty, or in the presence of mistakes or failures (Desai et al., 2012; Salomons et al., 2018). One of the novel contributions of the current paper lies in the observation that the degree of reliance on the help of a robot can be modulated, in turn, by the reliance of the robot itself on its human partner. Although in some contexts robots' vulnerability and susceptibility may generate positive emotional responses (Strohkorb Sebo et al., 2018), our results highlight how interaction with a susceptible robot may lead to detrimental consequences in terms of confidence in the robot's competence and, consequently, in individuals' reliance on its help and advice. This is particularly important if we consider that, in our task, the robot's performance was much higher than that of participants, meaning that its help would have been a valuable tool for behavioral improvement. These findings indicate that transparency is needed to correctly interpret behavior and capabilities of a robotic system and prevent underreliance and disuse from the user (Lee and See, 2004; Ososky et al., 2014; Parasuraman and Manzey, 2010; Wang et al., 2015; Złotowski et al., 2016). Moreover, these results are in line with research in collective and social decision-making showing that humans' susceptibility to information provided by others depends on the relative confidence of the interacting partners and on the possibility to communicate their level of confidence accurately to each other (Bahrami et al., 2010, 2012; Koriat, 2012).

We have shown that relational and normative effects of reciprocal social influence did emerge in the Robot group. In particular, results revealed that the observed *implicit* loss of confidence in the robot's competence (i.e., disuse of its feedback for learning purposes) was not accompanied by a decrease in the participants' *explicit* susceptibility to the robot, as overtly expressed during their final decisions (that were observed by the robot itself). A possible explanation of this effect is that participants were not willing to reveal a loss of confidence in a humanoid robot's competence that was relying heavily on the participants' estimates. Previous research on social influence among human peers (Mahmoodi et al., 2018; Zonca et al., 2021a, 2021b) has suggested that showing some level of susceptibility toward a partner may send a pro-social signal to the partner itself, since receiving consideration from peers is processed by the human brain as a positive reward (Hertz et al., 2017), whereas being ignored represents a negative reward (Eisenberger et al., 2003). In this context, taking into account the opinions of others (and not ignoring them) is perceived as a social norm (Mahmoodi et al., 2018). In line with this view, if the HRI assumes a value on the social level, it is possible that humans can be pleased, in some measure, by the influence exerted over the robot. In other words, individuals who have been recently pleased by the robot may feel the urge of behaving nicely with it or, at least, abstaining from revealing a loss of confidence in the robot's abilities, as commonly observed in human-human cooperative settings (Bartlett and DeSteno, 2006; Berns et al., 2010). Our findings are consistent with recent evidence revealing the emergence of pro-social attitudes toward robots in adults (Connolly et al., 2020; Kahn et al., 2015; Kühnlenz et al., 2018; Siegel et al., 2009) and children (Beran et al., 2011; Chernyak and Gary, 2016; Martin et al., 2020; Zaga et al., 2017). Moreover, they can provide a pro-social interpretation of recent results revealing over-trust with the instructions of faulty or unreliable robots (Aroyo et al., 2018, 2021; Robinette et al., 2016; Salem et al., 2015).

Nevertheless, we acknowledge that the normative effects of reciprocity observed in the Robot group in the current work are qualitatively and quantitatively different from those observed in HHI. Using a similar experimental design consisting of a dyadic interaction between a participant and an alleged *human* partner, Mahmoodi et al. (2018) have shown that participants' influence was significantly higher when the partner had shown high susceptibility (Susceptible condition) than when the partner had shown low susceptibility to the participant's judgments (Unsusceptible condition). These findings can be explained by a normative effect of reciprocity, whereas they are inconsistent with a decision-making model that takes into account only informational factors such as confidence, uncertainty, and information reliability. However, in our task, participants interacting with either a computer or a robot *did not* rely more on the partner's judgments in the Susceptible condition (rather, participants in both groups show a non-significant trend in the opposite direction). This result confirms that social influence in human-computer and human-robot scenarios is primarily modulated by the perception of the agent's competence and performance (Billings et al., 2012; Hancock et al., 2011, 2020; van den Brule et al., 2014; Wright et al., 2019). It is clear, and perhaps not surprising, that the social and normative value underlying human-robot relationships, similarly to that underlying general human-machine interactions, is qualitatively different from that observed in human peer relationships. This fundamental difference may be grounded in different factors. First, human participants may be less concerned about the well-being of a robot, since they do not believe that their behavior may cause negative consequences for the robot itself. Second, participants might not attribute social value and motives to the robot's behavior: for instance, the robot's susceptibility in the Reciprocal social influence task may be (mostly) interpreted as evidence of the robot's uncertainty rather than a pro-social signal.

Third, participants may not believe that their own behavior would have an impact on the behavior of the robot. In this regard, studies on human-human social influence (Mahmoodi et al., 2018; Zonca et al., 2021a) have suggested that reciprocity of social influence between human peers may emerge because of the desire to maintain their influence over a partner by controlling their own behavior. This phenomenon assumes that people tend to act pro-socially in response to others' pro-social acts because they believe that this may preserve their relational status and the associated reward (Campbell-Meiklejohn et al., 2010; Hertz et al., 2017; Izuma and Adolphs, 2013). However, results of the Final block, in which we manipulated participants' expectation of future interactions with either the computer or the robotic partner, did not reveal a clear effect of experimental group on the modulation of participants' susceptibility depending on the knowledge of future interactions with the partner. One hypothesis is that the normative phenomena underlying human-robot reciprocity, in our experimental setting, are not relevant enough to lead participants to use reciprocity instrumentally, in order to obtain consideration from the robot.

It must be noted that participants did not receive any information about the motives driving the behavior of the robot (and the computer). Therefore, participants' susceptibility expressed a genuine reaction to the perceived capabilities and motives underlying the behavior of their interacting partner. However, it is unclear how reciprocity (and, in general, pro-social behavior) can emerge when interacting with mechanical agents without transparent goals, motives, and desires. For instance, the emergence of a strong effect of reciprocity in HRI would require humans to know that robots are aware of social norms and may comply with them to achieve successful collaboration or please their human partner(s). In this sense, transparency about the purposes underlying the behavior of a robotic system may be crucial in promoting human-robot collaboration (Ososky et al., 2014). Our results indicate the need of more integrated models of HRI that treat the robot as an intentional agent with its own goals, motives, and desires (Bossi et al., 2020; Man and Damasio, 2019; Sciutti et al., 2013; Thellman et al., 2017; Wiese et al., 2017; Ziemke, 2020). Building an incorrect model of the capabilities and the purposes driving the behavior of the robotic system may determine its misuse or disuse (Ososky et al., 2013) or impede the establishment of successful human-robot collaboration.

Altogether, our results reveal that social influence in HRI does not follow the strict normative constraints observed in HHI (Mahmoodi et al., 2018; Zonca et al., 2021a, 2021b). Nonetheless, they reveal that the interaction with social robots is significantly different from that observed with simple computer algorithms, both on informational and social levels. These findings stress the importance of a bidirectional view of HRI, which does not focus exclusively on robots' appearance and capabilities but also focuses on the relational dynamics intervening during interaction. In this regard, we highlight that recent cognitive architectures have been paving the way for a more relational and reciprocal view of HRI, for instance, in the study of human-robot trust (Vinanzi et al., 2019, 2021). Robots should not just behave reliably but also possess the ability to correctly interpret the relational and social signals expressed by their human partner(s) and know how to react dynamically to preserve or improve collaboration. These abilities may be crucial in the design of robotic agents that could effectively act as collaborative companions in contexts such as healthcare (Robinson et al., 2014), rehabilitation (Kellmeyer et al., 2018), elderly people assistance (Frennert and Östlund, 2014), and education (Basoeki et al., 2013; Belpaeme et al., 2018).

## Limitations of the study

We acknowledge that the current work did non entail a physical interaction with the robot during the joint experimental tasks in the Robot group. On the one hand, this guaranteed complete comparability with the Computer group, ruling out potential confounds linked to implicit motor feedback arising from robot's movements, which would have affected the perception of the robot's competence. On the other hand, this experimental choice implies caution in the generalization of our findings to other types of (physical) HRI. Extensive research in HRI (e.g., Connolly et al., 2020; Kahn et al., 2015; Kühnlenz et al., 2018; Terzioğlu et al., 2020) has shown that the physical presence of a (social) robot with human-like behavior may trigger emotional and empathic reactions in human participants. We hypothesize that, in the presence of specific physical, motor, or verbal cues coming from a robot, the effects of reciprocity observed in the current work might be either amplified or reduced following pro-social or anti-social signals. In this regard, we also acknowledge that the modalities of the encounter with the robot (before starting the experimental session) may have affected our results. In particular, the meeting with iCub consisted of a simple, scripted speech of the robot and did not entail a free interaction with the robot with a dynamic dialogue or affective components. The encounter with iCub aimed at ensuring that the participant believed that the robot was an intentional agent that knew about the presence of the participant (i.e., by enabling head/gaze following) and the upcoming joint task. We did not create a particular social bound between the robot and the participant, since we aimed at studying the effect of the belief on the nature of the partner (robot or computer) in modulating the emergence of social influence effects with a perfectly controlled experimental setting. However, we believe that the *social* effects observed in the current study may be amplified or even canceled depending on the social attitude of the robot in the introductory phase, or by the use of a priming task involving social dimensions like cooperativeness or competitiveness. Future studies may explore the impact of socially relevant, embodied signals generated by social robots on reciprocal social influence during HRI.

Furthermore, our study of reciprocal human-robot social influence was applied to a functional, quantitative perceptual task. Researchers should be careful in generalizing our findings to experiments involving decisions or opinions on social issues. Recent evidence has shown an increased reliance on robots in functional rather than social tasks (Gaudiello et al., 2016; Ullman et al., 2021). Indeed, differences between the perceived robot's competence in functional and social tasks may have an impact on the emergence of reciprocal mechanisms of social influence in HRI.

## STAR★METHODS

Detailed methods are provided in the online version of this paper and include the following:

- KEY RESOURCES TABLE
- RESOURCE AVAILABILITY
  - ○ Lead contact
  - ○ Materials availability
  - ○ Data and code availability
- METHOD DETAILS
  - ○ Overview: participants and procedure
- INTRODUCING PARTICIPANTS TO THE HUMANOID ROBOT iCub
- TASKS DESCRIPTION
  - ○ Perceptual inference task
  - ○ Social influence task
  - ○ Reciprocal social influence task
- AGENTS' BEHAVIOR
  - ○ Agents' perceptual estimates
  - ○ Agents' final decisions in the reciprocal social influence task
- QUANTIFICATION AND STATISTICAL ANALYSIS

## SUPPLEMENTAL INFORMATION

Supplemental information can be found online at https://doi.org/10.1016/j.isci.2021.103424.

## ACKNOWLEDGMENTS

We gratefully acknowledge the financial support of the European Research Council (ERC Starting Grant 804388, wHiSPER). We would like to thank Fabio Vannucci for the insightful comments.

## AUTHOR CONTRIBUTIONS

J.Z., A.S., and A.F. designed the experimental protocol. J.Z. and A.F. programmed the experimental tasks. J.Z. and A.F. collected the data. J.Z. carried out the data analysis and wrote the manuscript. A.S. and A.F. provided suggestions for improving the manuscript.

## DECLARATION OF INTERESTS

The authors declare no competing interests.

## REFERENCES

Admoni, H., and Scassellati, B. (2017). Social eye gaze in human-robot interaction: a review. J. Hum. Robot Interact. *6*, 25–63. https://doi.org/10.5898/JHRI.6.1.

Aroyo, A.M., Pasquali, D., Kothig, A., Rea, F., Sandini, G., and Sciutti, A. (2021). Expectations vs. reality: unreliability and transparency in a treasure hunt game with iCub. IEEE Robot. Autom. Lett. *6*, 5681–5688. https://doi.org/10.1109/LRA.2021.3083465.

Aroyo, A.M., Rea, F., Sandini, G., and Sciutti, A. (2018). Trust and social engineering in human robot interaction: will a robot make you disclose sensitive information, conform to its recommendations or gamble? IEEE Robot. Autom. Lett. *3*, 3701–3708. https://doi.org/10.1109/LRA.2018.2856272.

Asch, S.E. (1951). Effects of group pressure upon the modification and distortion of judgment. In Groups, Leadership and Men, H. Guetzkow, ed. (Carnegie Press), pp. 177–190.

Bahrami, B., Olsen, K., Latham, P.E., Roepstorff, A., Rees, G., and Frith, C.D. (2010). Optimally interacting minds. Science *329*, 1081–1085. https://doi.org/10.1126/science.1185718.

Bahrami, B., Olsen, K., Bang, D., Roepstorff, A., Rees, G., and Frith, C. (2012). What failure in collective decision-making tells us about metacognition. Philos. Trans. R. Soc. B Biol. Sci. *367*, 1350–1365. https://doi.org/10.1098/rstb.2011.0420.

Baraglia, J., Cakmak, M., Nagai, Y., Rao, R.P., and Asada, M. (2017). Efficient human-robot collaboration: when should a robot take initiative? Int. J. Rob. Res. *36*, 563–579. https://doi.org/10.1177/0278364916688253.

Barta, Z., McNamara, J.M., Huszar, D.B., and Taborsky, M. (2011). Cooperation among non-relatives evolves by state-dependent generalized reciprocity. Proc. R. Soc. B *278*, 843–848. https://doi.org/10.1098/rspb.2010.163.

Bartlett, M.Y., and DeSteno, D. (2006). Gratitude and prosocial behavior. Psychol. Sci. *17*, 319–325. https://doi.org/10.1111/j.1467-9280.2006.01705.x.

Basoeki, F., Dalla Libera, F., Menegatti, E., and Moro, M. (2013). Robots in education: new trends and challenges from the Japanese market. Themes Sci. Technol. Educ. *6*, 51–62.

Behrens, T.E., Hunt, L.T., Woolrich, M.W., and Rushworth, M.F. (2008). Associative learning of social value. Nature *456*, 245–249. https://doi.org/10.1038/nature07538.

Belpaeme, T., Kennedy, J., Ramachandran, A., Scassellati, B., and Tanaka, F. (2018). Social robots for education: a review. Sci. Robotics *3*, eaat5954. https://doi.org/10.1126/scirobotics.aat5954.

Beran, T.N., Ramirez-Serrano, A., Kuzyk, R., Nugent, S., and Fior, M. (2011). Would children help a robot in need? Int. J. Soc. Robot. *3*, 83–93. https://doi.org/10.1007/s12369-010-0074-7.

Berns, G.S., Capra, C.M., Moore, S., and Noussair, C. (2010). Neural mechanisms of the influence of popularity on adolescent ratings of

music. Neuroimage *49*, 2687–2696. https://doi.org/10.1016/j.neuroimage.2009.10.070.

Billings, D.R., Schaefer, K.E., Chen, J.Y., and Hancock, P.A. (2012). Human-robot interaction: developing trust in robots. In Proc. of ACM/IEEE Int. Conf. Human-Robot Interact., pp. 109–110. https://doi.org/10.1145/2157689.2157709.

Bonaccio, S., and Dalal, R.L. (2010). Evaluating advisors: a policy-capturing study under conditions of complete and missing information. J. Behav. Decis. Mak. *23*, 227–249. https://doi.org/10.1002/bdm.649.

Boorman, E.D., O'Doherty, J.P., Adolphs, R., and Rangel, A. (2013). The behavioral and neural mechanisms underlying the tracking of expertise. Neuron *80*, 1558–1571. https://doi.org/10.1016/j.neuron.2013.10.024.

Bossi, F., Willemse, C., Cavazza, J., Marchesi, S., Murino, V., and Wykowska, A. (2020). The human brain reveals resting state activity patterns that are predictive of biases in attitudes toward robots. Sci. Robot. *5*, eabb6652. https://doi.org/10.1126/scirobotics.abb6652.

Campbell-Meiklejohn, D.K., Bach, D.R., Roepstorff, A., Dolan, R.J., and Frith, C.D. (2010). How the opinion of others affects our valuation of objects. Curr. Biol. *20*, 1165–1170. https://doi.org/10.1016/j.cub.2010.04.055.

Chernyak, N., and Gary, H.E. (2016). Children's cognitive and behavioral reactions to an autonomous versus controlled social robot dog. Early Educ. Dev. *27*, 1175–1189. https://doi.org/10.1080/10409289.2016.1158611.

Cialdini, R.B., and Goldstein, N.J. (2014). Social influence: compliance and conformity. Annu. Rev. Psychol. *55*, 591–621. https://doi.org/10.1146/annurev.psych.55.090902.142015.

Cicchini, G.M., Arrighi, R., Cecchetti, L., Giusti, M., and Burr, D.C. (2012). Optimal encoding of interval timing in expert percussionists. J. Neurosci. *32*, 1056–1060. https://doi.org/10.1523/JNEUROSCI.3411-11.2012.

Claidière, N., and Whiten, A. (2012). Integrating the study of conformity and culture in humans and nonhuman animals. Psychol. Bull. *138*, 126–145. https://doi.org/10.1037/a0025868.

Cohen, B. (2008). Explaining Psychological Statistics (John Wiley and Sons).

Connolly, J., Mocz, V., Salomons, N., Valdez, J., Tsoi, N., Scassellati, B., and Vázquez, M. (2020). Prompting prosocial human interventions in response to robot mistreatment. In Proc. ACM/IEEE Int. Conf. Human-Robot Interact., pp. 211–220. https://doi.org/10.1145/3319502.3374781.

De Martino, B., Bobadilla-Suarez, S., Nouguchi, T., Sharot, T., and Love, B.C. (2017). Social information is integrated into value and confidence judgments according to its reliability. J. Neurosci. *37*, 6066–6074. https://doi.org/10.1523/JNEUROSCI.3880-16.2017.

Desai, M., Medvedev, M., Vázquez, M., McSheehy, S., Gadea-Omelchenko, S., Bruggeman, C., Steinfeld, A., and Yanco, H.

(2012). Effects of changing reliability on trust of robot systems. In Proc. ACM/IEEE Int. Conf. Human-Robot Interact., pp. 73–80. https://doi.org/10.1145/2157689.2157702.

Eisenberger, N.I., Lieberman, M.D., and Williams, K.D. (2003). Does rejection hurt? An fMRI study of social exclusion. Science *302*, 290–292. https://doi.org/10.1126/science.1089134.

Flook, R., Shrinah, A., Wijnen, L., Eder, K., Melhuish, C., and Lemaignan, S. (2019). On the impact of different types of errors on trust in human-robot interaction: are laboratory-based HRI experiments trustworthy? Interact. Stud. *20*, 455–486. https://doi.org/10.1075/is.18067.flo.

Frennert, S., and Östlund, B. (2014). Seven matters of concern of social robots and older people. Int. J. Soc. Robot *6*, 299–310. https://doi.org/10.1007/s12369-013-0225-8.

Fritz, C.O., Morris, P.E., and Richler, J.J. (2012). Effect size estimates: current use, calculations, and interpretation. J. Exp. Psychol. Gen. *141*, 2. https://doi.org/10.1037/a0026092.

Gaudiello, I., Zibetti, E., Lefort, S., Chetouani, M., and Ivaldi, S. (2016). Trust as indicator of robot functional and social acceptance. An experimental study on user conformation to iCub answers. Comput. Hum. Behav. *61*, 633–655. https://doi.org/10.1016/j.chb.2016.03.057.

Gouldner, A.W. (1960). The norm of reciprocity: a preliminary statement. Am. Sociol. Rev. *25*, 161–178. https://doi.org/10.2307/2092623.

Hancock, P.A., Billings, D.R., Schaefer, K.E., Chen, J.Y., De Visser, E.J., and Parasuraman, R. (2011). A meta-analysis of factors affecting trust in human-robot interaction. Hum. Factors *53*, 517–527. https://doi.org/10.1177/0018720811417254.

Hancock, P.A., Kessler, T.T., Kaplan, A.D., Brill, J.C., and Szalma, J.L. (2020). Evolving trust in robots: specification through sequential and comparative meta-analyses. Hum. Factors *53*, 517–527. https://doi.org/10.1177/0018720820922080.

Hertz, U., Palminteri, S., Brunetti, S., Olesen, C., Frith, C.D., and Bahrami, B. (2017). Neural computations underpinning the strategic management of influence in advice giving. Nat. Commun. *8*, 1–12. https://doi.org/10.1038/s41467-017-02314-5.

Hertz, N., and Wiese, E. (2018). Under pressure: examining social conformity with computer and robot groups. Hum. Factors *60*, 1207–1218. https://doi.org/10.1177/0018720818788473.

Hollingworth, H.L. (1910). The central tendency of judgment. J. Philos. Psychol. Sci. Methods *7*, 461–469.

Izuma, K., and Adolphs, R. (2013). Social manipulation of preference in the human brain. Neuron *78*, 563–573. https://doi.org/10.1016/j.neuron.2013.03.023.

Jazayeri, M., and Shadlen, M.N. (2010). Temporal context calibrates interval timing. Nat. Neurosci. *13*, 1020. https://doi.org/10.1038/nn.2590.

Kahn, P.H., Jr., Kanda, T., Ishiguro, H., Gill, B.T., Shen, S., Gary, H.E., and Ruckert, J.H. (2015). Will people keep the secret of a humanoid robot? Psychological intimacy in HRI. In Proc. ACM/IEEE Int. Conf. Human-robot Interact., pp. 173–180. https://doi.org/10.1145/2696454.2696486.

Kaniarasu, P., Steinfeld, A., Desai, M., and Yanco, H. (2013). Robot confidence and trust alignment. In Proc. ACM/IEEE Int. Conf. Human-robot Interact., pp. 155–156. https://doi.org/10.1109/HRI.2013.6483548.

Kellmeyer, P., Mueller, O., Feingold-Polak, R., and Levy-Tzedek, S. (2018). Social robots in rehabilitation: a question of trust. Sci. Robot. 3, eaat1587. https://doi.org/10.1126/scirobotics.aat1587.

Koriat, A. (2012). When are two heads better than one and why? Science 336, 360–362. https://doi.org/10.1126/science.1216549.

Kühnlenz, B., Kühnlenz, K., Busse, F., Förtsch, P., and Wolf, M. (2018). Effect of explicit emotional adaptation on prosocial behavior of humans towards robots depends on prior robot experience. In Proc. of IEEE Int. Symp. on Robot and Human Interactive Communication, pp. 275–281. https://doi.org/10.1109/ROMAN.2018.8525515.

Lee, J.D., and See, K.A. (2004). Trust in automation: designing for appropriate reliance. Hum. Factors 46, 50–80. https://doi.org/10.1518/hfes.46.1.50_30392.

Lussier, B., Gallien, M., and Guiochet, J. (2007). Fault tolerant planning for critical robots. In Proc. of the 37th Annual IEEE/IFIP International Conference on Dependable Systems and Networks. https://doi.org/10.1109/DSN.2007.50.

Mahmoodi, A., Bahrami, B., and Mehring, C. (2018). Reciprocity of social influence. Nat. Commun. 9, 1–9. https://doi.org/10.1038/s41467-018-04925-y.

Man, K., and Damasio, A. (2019). Homeostasis and soft robotics in the design of feeling machines. Nat. Mach. Intell. 1, 446–452. https://doi.org/10.1038/s42256-019-0103-7.

Martin, D.U., Perry, C., MacIntyre, M.I., Varcoe, L., Pedell, S., and Kaufman, J. (2020). Investigating the nature of children's altruism using a social humanoid robot. Comput. Hum. Behav. 104, 106149. https://doi.org/10.1016/j.chb.2019.09.025.

Metta, G., Fitzpatrick, P., and Natale, L. (2006). YARP: yet another robot platform. Int. J. Adv. Robot. Syst. 3, 043–048. https://doi.org/10.5772/5761.

Metta, G., Natale, L., Nori, F., Sandini, G., Vernon, D., Fadiga, L., von Hofsten, C., Rosander, K., Lopes, M., Santos-Victor, J., et al. (2010). The iCub humanoid robot: an open-systems platform for research in cognitive development. Neural Netw. 23, 1125–1134. https://doi.org/10.1016/j.neunet.2010.08.010.

Metta, G., Sandini, G., Vernon, D., Natale, L., and Nori, F. (2008). The iCub humanoid robot: an open platform for research in embodied cognition. In Proc. of the 8th Workshop on Performance Metrics for Intelligent Systems,

pp. 50–56. https://doi.org/10.1145/1774674.1774683.

Morin, O., Jacquet, P.O., Vaesen, K., and Acerbi, A. (2021). Social information use and social information waste. Phil. Trans. R. Soc. B 376, 20200052. https://doi.org/10.1098/rstb.2020.0052.

Nowak, M.A. (2006). Five rules for the evolution of cooperation. Science 314, 1560–1563. https://doi.org/10.1126/science.1133755.

Nowak, M.A., and Sigmund, K. (1998). Evolution of indirect reciprocity by image scoring. Nature 393, 573–577. https://doi.org/10.1038/31225.

Ohtsuki, H., and Iwasa, Y. (2004). How should we define goodness?—Reputation dynamics in indirect reciprocity. J. Theor. Biol. 231, 107–120. https://doi.org/10.1016/j.jtbi.2004.06.005.

Oliveira, R., Arriaga, P., Santos, F.P., Mascarenhas, S., and Paiva, A. (2021). Towards prosocial design: a scoping review of the use of robots and virtual agents to trigger prosocial behaviour. Comput. Hum. Behav. 114, 106547. https://doi.org/10.1016/j.chb.2020.106547.

Ososky, S., Sanders, T., Jentsch, F., Hancock, P., and Chen, J.Y. (2014). Determinants of system transparency and its influence on trust in and reliance on unmanned robotic systems. In Unmanned Systems Technology XVI, 9084 (International Society for Optics and Photonics), p. 90840E. https://doi.org/10.1117/12.2050062.

Ososky, S., Schuster, D., Phillips, E., and Jentsch, F.G. (2013). Building appropriate trust in human-robot teams. In 2013 AAAI Spring Symposium Series.

Parasuraman, R., and Manzey, D.H. (2010). Complacency and bias in human use of automation: an attentional integration. Hum. Factors 52, 381–410. https://doi.org/10.1177/0018720810376055.

Park, S.A., Goïame, S., O'Connor, D.A., and Dreher, J.C. (2017). Integration of individual and social information for decision-making in groups of different sizes. PLoS Biol. 15, e2001958. https://doi.org/10.1371/journal.pbio.2001958.

Petzschner, F.H., Glasauer, S., and Stephan, K.E. (2015). A Bayesian perspective on magnitude estimation. Trends Cogn. Sci. 19, 285–293. https://doi.org/10.1016/j.tics.2015.03.002.

Pfeiffer, T., Rutte, C., Killingback, T., Taborsky, M., and Bonhoeffer, S. (2005). Evolution of cooperation by generalized reciprocity. Proc. R. Soc. B 272, 1115–1120. https://doi.org/10.1098/rspb.2004.2988.

Robinette, P., Li, W., Allen, R., Howard, A.M., and Wagner, A.R. (2016). Overtrust of robots in emergency evacuation scenarios. In Proc. ACM/IEEE Int. Conf. Human-Robot Interact., pp. 101–108. https://doi.org/10.1109/HRI.2016.7451740.

Robinson, H., MacDonald, B., and Broadbent, E. (2014). The role of healthcare robots for older people at home: a review. Int. J. Soc. Robot. 6, 575–591. https://doi.org/10.1007/s12369-014-0242-2.

Rossi, A., Dautenhahn, K., Koay, K.L., and Walters, M.L. (2017). How the timing and magnitude of robot errors influence peoples' trust of robots in an emergency scenario. In Social Robotics. ICSR 2017. Lecture Notes in Computer Science, 10652, A. Kheddar, et al., eds. (Springer), pp. 44–52. https://doi.org/10.1007/978-3-319-70022-9_5.

Salem, M., Lakatos, G., Amirabdollahian, F., and Dautenhahn, K. (2015). Would you trust a (faulty) robot? Effects of error, task type and personality on human-robot cooperation and trust. In Proc. of ACM/IEEE Int. Conf. Human-Robot Interact., pp. 1–8.

Salomons, N., van der Linden, M., Strohkorb, S., and Scassellati, B. (2018). Humans conform to robots: disambiguating trust, truth, and conformity. In Proc. of ACM/IEEE Int. Conf. Human-Robot Interact., pp. 187–195. https://doi.org/10.1145/3171221.3171282.

Sanders, T., Oleson, K.E., Billings, D.R., Chen, J.Y., and Hancock, P.A. (2011). A model of human-robot trust: theoretical model development. Proc. Hum. Factors Ergon. Soc. Annu. Meet. 55, 1432–1436. https://doi.org/10.1177/1071181311551298.

Sandini, G., and Sciutti, A. (2018). Humane robots—from robots with a humanoid body to robots with an anthropomorphic mind. ACM Trans. Hum. Robot Interact. 7. https://doi.org/10.1145/3208954.

Sciutti, A., Bisio, A., Nori, F., Metta, G., Fadiga, L., and Sandini, G. (2013). Robots can be perceived as goal-oriented agents. Interact. Stud. 14, 329–350. https://doi.org/10.1075/is.14.3.02sci.

Sciutti, A., Burr, D., Saracco, A., Sandini, G., and Gori, M. (2014). Development of context dependency in human space perception. Exp. Brain Res. 232, 3965–3976. https://doi.org/10.1007/s00221-014-4021-y.

Siegel, M., Breazeal, C., and Norton, M.I. (2009). Persuasive robotics: the influence of robot gender on human behavior. In IEEE/RSJ Int. Conf. Intell. Robots Syst., pp. 2563–2568. https://doi.org/10.1109/IROS.2009.5354116.

Sniezek, J.A., and Van Swol, L.M. (2001). Trust, confidence, and expertise in a judge–advisor system. Organ. Behav. Hum. Decis. Process. 84, 288–307. https://doi.org/10.1006/obhd.2000.2926.

Sniezek, J.A., Schrah, G.E., and Dalal, R.S. (2004). Improving judgment with prepaid expert advice. J. Behav. Decis. Mak. 17, 173–190. https://doi.org/10.1002/bdm.468.

Strohkorb, S., and Scassellati, B. (2016). Promoting collaboration with social robots. In Proc. ACM/IEEE Int. Conf. Human-robot Interact., pp. 639–640. https://doi.org/10.1109/HRI.2016.7451895.

Strohkorb Sebo, S., Traeger, M., Jung, M., and Scassellati, B. (2018). The ripple effects of vulnerability: the effects of a robot's vulnerable behavior on trust in human-robot teams. In Proc. ACM/IEEE Int. Conf. Human-robot Interact., pp. 178–186. https://doi.org/10.1145/3171221.3171275.

Terzioğlu, Y., Mutlu, B., and Şahin, E. (2020). Designing social cues for collaborative robots:

the role of gaze and breathing in human-robot collaboration. In Proc. of ACM/IEEE Int. Conf. Human-Robot Interact., pp. 343–357. https://doi.org/10.1145/3319502.3374829.

Thellman, S., Silvervarg, A., and Ziemke, T. (2017). Folk-psychological interpretation of human vs. humanoid robot behavior: exploring the intentional stance toward robots. Front. Psychol. 8, 1962. https://doi.org/10.3389/fpsyg.2017.01962.

Tsagarakis, N.G., Metta, G., Sandini, G., Vernon, D., Beira, R., Becchi, F., Righetti, L., Santos-Victor, J., Ijspeert, A.J., Carrozza, M.C., and Caldwell, D.G. (2007). iCub: the design and realization of an open humanoid platform for cognitive and neuroscience research. Adv. Robot. 21, 1151–1175. https://doi.org/10.1163/156855307781389419.

Ullman, D., Aladia, S., and Malle, B.F. (2021). Challenges and opportunities for replication science in HRI: a case study in human-robot trust. In Proc. of ACM/IEEE Int. Conf. Human-Robot Interact., pp. 110–118. https://doi.org/10.1145/3434073.3444652.

van den Brule, R., Dotsch, R., Bijlstra, G., Wigboldus, D.H., and Haselager, P. (2014). Do robot performance and behavioral style affect human trust? Int. J. Soc. Robot. 6, 519–531. https://doi.org/10.1007/s12369-014-0231-5.

Vélez, N., and Gweon, H. (2019). Integrating incomplete information with imperfect advice. Top. Cogn. Sci. 11, 299–315. https://doi.org/10.1111/tops.12388.

Vinanzi, S., Cangelosi, A., and Goerick, C. (2021). The collaborative mind: intention reading and trust in human-robot interaction. iScience 24, 102130. https://doi.org/10.1016/j.isci.2021.102130.

Vinanzi, S., Patacchiola, M., Chella, A., and Cangelosi, A. (2019). Would a robot trust you? Developmental robotics model of trust and theory of mind. Philos. Trans. R. Soc. B 374, 20180032. https://doi.org/10.1098/rstb.2018.0032.

Wang, N., Pynadath, D.V., Hill, S.G., and Ground, A.P. (2015). Building trust in a human-robot team with automatically generated explanations. In Proc. Interservice/Industry Training, Simulation Education Conf., 15315Proc. Interservice/Industry Training, Simulation Education Conf., pp. 1–12.

Wiese, E., Metta, G., and Wykowska, A. (2017). Robots as intentional agents: using neuroscientific methods to make robots appear more social. Front. Psychol. 8, 1663. https://doi.org/10.3389/fpsyg.2017.01663.

Wright, J.L., Chen, J.Y., and Lakhmani, S.G. (2019). Agent transparency and reliability in human–robot interaction: the influence on user confidence and perceived reliability. IEEE Trans. Hum. Mach. Syst. 50, 254–263. https://doi.org/10.1109/THMS.2019.2925717.

Xu, A., and Dudek, G. (2016). Maintaining efficient collaboration with trust-seeking robots. In 2016 IEEE/RSJ Int. Conf. Intell. Robots Syst., pp. 3312–3319. https://doi.org/10.1109/IROS.2016.7759510.

Yaniv, I. (2004). Receiving other people's advice: influence and benefit. Organ. Behav. Hum. Decis. Process. 93, 1–13. https://doi.org/10.1016/j.obhdp.2003.08.002.

Yaniv, I., and Kleinberger, E. (2000). Advice taking in decision making: egocentric discounting and reputation formation. Organ. Behav. Hum. Decis. Process. 83, 260–281. https://doi.org/10.1006/obhd.2000.2909.

Zaga, C., Moreno, A., and Evers, V. (2017). Gotta hatch'em all!: robot-supported cooperation in interactive playgrounds. In Companion of the 2017 ACM Conference on Computer Supported Cooperative Work and Social Computing, pp. 347–350. https://doi.org/10.1145/3022198.3026355.

Ziemke, T. (2020). Understanding robots. Sci. Robot. 5, eabe2987. https://doi.org/10.1126/scirobotics.abe2987.

Złotowski, J., Sumioka, H., Nishio, S., Glas, D.F., Bartneck, C., and Ishiguro, H. (2016). Appearance of a robot affects the impact of its behaviour on perceived trustworthiness and empathy. Paladyn, J. Behav. Robot. 7. https://doi.org/10.1515/pjbr-2016-0005.

Zonca, J., Folsø, A., and Sciutti, A. (2021a). Dynamic modulation of social influence by indirect reciprocity. Sci. Rep. 11, 11104. https://doi.org/10.1038/s41598-021-90656-y.

Zonca, J., Folsø, A., and Sciutti, A. (2021b). I'm not a little kid anymore! Reciprocal social influence in child-adult interaction. R. Soc. Open Sci. 8, 202124. https://doi.org/10.1098/rsos.202124.

## STAR★METHODS

### KEY RESOURCES TABLE

| REAGENT or RESOURCE | SOURCE | IDENTIFIER |
| --- | --- | --- |
| Deposited Data | | |
| Processed datasets and figure codes | Open Science Framework repository | https://osf.io/3yhua/?view_only=54c5ed45f41c4ed1955ecfba85e6ccfa |
| Software and Algorithms | | |
| Robot's custom script | Open Science Framework repository | https://osf.io/3yhua/?view_only=54c5ed45f41c4ed1955ecfba85e6ccfa |
| YARP | YARP: Yet Another Robot Platform | https://www.yarp.it/latest/ |
| Other | | |
| iCub 2 humanoid robot | Istituto Italiano di Tecnologia | https://icub.iit.it/ |

### RESOURCE AVAILABILITY

#### Lead contact

Further information and requests for resources should be directed to and will be fulfilled by the Lead Contact, Joshua Zonca (joshua.zonca@iit.it).

#### Materials availability

No materials were newly generated for this paper.

#### Data and code availability

- Datasets supporting analyses and figures included in the current study are available in a dedicated OSF repository. Accession numbers are listed in the key resources table.
- Codes used to implement the behavior of the robot, generate figures and support results are available in a dedicated OSF repository. Accession numbers are listed in the key resources table.

### METHOD DETAILS

#### Overview: participants and procedure

We collected data from 50 participants (22 females, mean age: 34.96, SD: 13.13). All participants completed the entire experimental paradigm, which included three different tasks that were performed in this exact order: Perceptual inference task, Social influence task and Reciprocal social influence task. Half of the participants were assigned to the Computer group, whereas the other half were assigned to the Robot group. The difference between the two groups lies in the participants' belief about the nature of the interacting partner in both the Social influence task and the Reciprocal social influence task: a humanoid robot iCub (Robot group) or a computer (Computer group). In fact, feedback concerning the partner's behavior in both groups was controlled by the same computer algorithms. Participants in the Robot group could not see their robotic partner while performing the tasks, but had the possibility to meet it before starting the experimental tasks (see the next paragraph "Introducing participants to the humanoid robot iCub"). Afterwards, participants in the Robot group were conducted in a new room to perform the experimental tasks. Participants in the Computer group underwent the same experimental paradigm as the one assigned to the Robot condition, but they did not meet the robot before the start of the experimental protocol and they were simply told they would interact with a computer. In both conditions, the experimental paradigm was carried out in a dimly lit room, to ensure an optimal visibility of the stimuli on the screen. Participants seated in front of a wide touch-screen tablet (43.69 × 24.07 cm), at a distance that allowed participants to see the visual stimuli, receive feedback from their partner and reach the screen to make decisions. In order to allow participants to respond with high spatial accuracy, we provided a touch-pen with an ultrathin tip. Before beginning the experimental tasks, written instructions were given and participants were allowed to ask questions.

Participants were told that their reimbursement would be calculated based on their performance and, in particular, on the accuracy of both their initial and final estimates (i.e., final decisions) in both tasks. The accuracy of the partner was not supposed to have an impact on participants' outcomes, and vice versa. However, everyone received the maximum amount (15 euros) at the end of the experiment, following the guidelines of the Italian Institute of Technology and the local ethics committee concerning the application of a fair reimbursement for voluntary participation in experimental research. The final debriefing revealed that all participants, during the experiment, believed that their final reimbursement would be affected by their performance. Eventually, we extensively debriefed participants about the experimental procedures, the reasons underlying the modality of reimbursement and the goals of our research, in accordance with the relevant ethical guidelines. The study was approved by the local ethics committee (Azienda Sanitaria Locale Genovese N.3, protocol: IIT_wHiSPER) and all participants gave informed consent.

## INTRODUCING PARTICIPANTS TO THE HUMANOID ROBOT ICUB

In order to investigate the processes of reciprocal social influence during an interaction with a social robot, we needed a robot that could exhibit human-like and social-like behavior. Therefore, we chose to use the humanoid robot iCub, which is an open source humanoid robot for research in embodied cognition and artificial intelligence (Metta et al., 2008, 2010).

The robot possesses 53 actuated degrees of freedom that permit human-like movement of the head, arms, hands, waist and legs. It is endowed with sensors and actuators allowing it to produce controlled, fine-grained actions and direct its attention towards objects and individuals. It also has LEDs mounted behind the face panel, to represent mouth and eyebrows, which enable it to produce facial expressions and simulate lip motion during speech. Thanks to such features, iCub is capable of showing human-like appearance and behavior (Metta et al., 2008; Tsagarakis et al., 2007) and be perceived as an intentional agent (Bossi et al., 2020; Sciutti et al., 2013; Thellman et al., 2017; Wiese et al., 2017; Ziemke, 2020) that is aware of the surrounding environment and autonomously generates actions to fulfill specific goals. For the purposes of the current experiment, we aimed at conveying the impression that the robot could 1) act as a social and intentional agent, which was aware of the presence of the participant and knew about the upcoming joint experiment and 2) physically perform the same task that participants would face in their experimental session.

To accomplish these goals, participants in the Robot group had a short meeting with iCub before starting the experiment. One experimenter accompanied the participant in the room with iCub, while another experimenter controlled the robot from the sidelines. The robot performed a series of predetermined actions through a custom-made script (see key resource table) running in the YARP environment (Metta et al., 2006). The researcher that controlled the robot on the sidelines managed the timing of the robot actions to simulate a natural interaction with the human participant. Before the arrival of the participant, iCub was placed in front of a touch-screen tablet. The tablet was identical to the one that participants would use during their experimental session. Once the participant had been introduced in the room, iCub turned toward them saying hello with its voice and waving its hand. The participant was conducted in front of the robot, so that iCub could track their face and direct its attention towards them. Then the robot introduced itself and informed the participant that they would play a game together, while continuing to look at the participant and follow their head movements. These actions aimed at signaling to the participant that iCub was aware of their presence and knew that would interact with the participant in the upcoming experiment. Eventually, iCub said goodbye to the participant and turned towards the tablet, announcing that it was ready to play. In order to give the impression that iCub was able to observe stimuli on the touch-screen tablet and reach it with its hand to perform the task, it also leaned forward and moved its right arm and hand in the direction of the tablet, pointing in the direction of the screen with its right index, as if it was ready to touch the screen. At this point, the participant was accompanied in another room to start the experimental session.

## TASKS DESCRIPTION

### Perceptual inference task

In each trial (Figure 1A), participants were presented with two consecutive light flashes (red disks of 0.57 cm of diameter, duration 200 ms) appearing on a visible horizontal white line crossing the whole screen at its central height. The first disk was positioned at a variable distance from the left border of the screen (0.6–6.6 cm). After its disappearance and an inter-stimulus interval of 200 ms, a second disk appeared at a variable distance to its right. We defined the distance between the first and the second disk as the target stimulus length (s). The target stimulus length (s) was randomly selected from 11 different sample distances (min: 8 cm, max: 16 cm, step: 0.8 cm). Participants were then asked to touch a point on the visible line, to the right of the second disk, in order to reproduce a segment (connecting the second and the third disk) matching the target stimulus length. Right after the touch of the screen, a third red disk appeared in the selected position. No feedback about the accuracy of the response was provided. The task consisted of 66 trials after which participants were asked to evaluate from 1 to 10 the accuracy of their perceptual estimates.

### Social influence task

At the beginning of each trial, participants made perceptual inferences in the same way as in the Perceptual inference task. The position of their selection was then marked with a vertical red line and the word YOU. Participants were told that, during this interval, the other agent (computer in the Computer group or robot in the Robot group) would see the same stimulus and would estimate its length. Then the other agent's estimate was shown along with the word PC or ICUB (depending on the experimental group) in blue (Figure 2). Importantly, the partner's feedback was always shown *after* the participant's decision (2.50–2.75 seconds after the participant's response) to prevent participants from being influenced by the estimate of their partner in their estimation process. Participants were informed about the presence of a delay in the appearance of the partner's feedback, in order to prevent participants to make inferences about the accuracy of the partner's responses based on its reaction times. Participants were not informed about the actual magnitude of this delay, which was chosen to appear credible also in the Robot condition, where the partner was likely to require some seconds to make a physical response on the touch-screen tablet.

Right after the partner's estimate, participants were asked to make a final decision by choosing any position between their previous response and the agent's response. In this way, participants' final decisions expressed the relative weight assigned to the judgment of the two interacting partners. However, participants were simply instructed to be as accurate as possible in both decisions and told that their accuracy in both decisions would affect their final score and reimbursement. After the participant's final decision, a green dot and a vertical green line with the word FINAL appeared in the position selected by the participant. Participants were told that their partner could see the position of both their initial perceptual estimates and their final decisions. The task consisted of 66 trials divided in three blocks by two brief pauses. The position of all the three responses (participant's estimate, partner's estimate and final decision) remained on screen for 1 s.

At the end of the task, participants evaluated (1–10) their own and the other agent's accuracy in the perceptual estimates (ignoring the perceived accuracy of the final decision).

### Reciprocal social influence task

The Reciprocal social influence task (Figure 2) was similar to the Social influence task, with the difference that, in half of the trials, the final decision was taken by the counterpart (computer or robot). The task consisted of two types of alternating turns, which prescribed the identity of the agent who would have taken the final decision. In *decision turns*, the participant made their perceptual estimate, then observed the response of the partner and eventually the participant had to take the final decision, as in the previous task. Participants were told that the other agent could observe both their perceptual inferences and final decisions. In *observation turns*, the participant made their estimate, then observed that of the partner and eventually observed the final decision made by the partner, which occurred 3.75–4.50 seconds after the appearance of the agent's original estimate. In the observation turns, participants could just observe the counterpart's final decisions and could not modify their estimates themselves. After feedback on the partner's final decision, the position of every response was kept visible on the screen for 1 s. Participants were instructed to be as accurate as possible in all the decisions they personally had to make, including both perceptual estimates and final decisions, and that the accuracy of all responses would contribute to their final score and their reimbursement.

The partner's final decisions were manipulated during the task in order to express two different types of behavior: in the *Susceptible* condition, the agent (computer or robot) was highly influenced by the participant's response, whereas in the *Unsusceptible* condition the agent tended to confirm its own first response and was much less influenced by the participant's response (for a detailed description of these algorithms, see the next paragraph "Agents' behavior"). Participants were not informed about the existence of these two different conditions. The presentation of Susceptible and Unsusceptible conditions followed a within-subject block design. The order of presentation of the two different blocks was counterbalanced across participants. Each of the two blocks consisted of 66 trials (33 decision turns and 33 observation turns), divided in three sub-blocks by two brief pauses.

Between the two main experimental blocks (Susceptible and Unsusceptible), we introduced a Transition block consisting of 22 trials characterized by a smooth transition in the behavior of the algorithmic agent in order to make the observed behavioral change appear more natural and gradual (see the paragraph "transition and final blocks" of the STAR Methods).

At the end of these 154 trials (Susceptible, Transition and Unsusceptible block), we added 22 trials characterized by a different alternation of turns (Final block): first participants faced 11 observation turns in a row, and eventually they underwent 11 decision turns in a row. Participants were informed about this change in the order of turns only at the beginning of the Final block and were informed that the experiment would finish after the 11 decision turns. We introduced the Final block to test whether participants would have changed their level of susceptibility to the partner if they did not expect future interactions with it.

The task consisted of a total of 176 trials. Participants were told that for the entire length of the task, including the Final block, their partner could observe both their initial and final responses. During the task there were pauses every 22 trials. In three of these pauses, participants were asked to evaluate from 1 to 10 their own and the other agent's accuracy in terms of perceptual estimation, without considering the final decisions. In addition they were asked to estimate how much the other agent would value the participants' accuracy (from 1 to 10). Specifically, participants were asked to rate accuracies once during the Susceptible block (in the last pause of the same block, after 44 susceptible trials), once right after the Transition block, and once during the Unsusceptible block (in the last pause of the block, after 44 unsusceptible trials).

## AGENTS' BEHAVIOR

### Agents' perceptual estimates

The simulated perceptual estimates of the interacting agent (computer or robot) were based on the same probabilistic algorithm. In each trial, the position of its perceptual estimate was randomly chosen from a Gaussian distribution centered at the correct response (SD: 1.52 cm). The algorithm was not characterized by central tendency effects, since the distribution of its perceptual estimates was centered on the correct response. The standard deviation of the response error distribution was chosen to maintain a balance between variability, credibility and accuracy of response. Indeed, the standard deviation of the response distribution of the algorithm was set to be 25% lower than the observed standard deviation of participants' perceptual inferences, as estimated in a pilot study. Our goal was to prevent participants from recurrently observing extremely high discrepancies between the two responses, which would have highly affected the perceived reliability of their partner. Nonetheless, we also considered the possibility that few participants could be extremely accurate in their perceptual estimates. In this scenario, the participant and partner would have selected close responses very often, impeding to observe variability in participants' final estimates. Therefore, in half of the instances in which the algorithm's sampled estimate was rather close to that of the participant (i.e., d < 0.5 cm), the algorithm re-sampled a new estimate from the distribution (i.e., until d > 0.5 cm).

This response distribution was used for all perceptual estimates in the Perceptual inference task, in the Social influence task and in the Reciprocal social influence task.

### Agents' final decisions in the reciprocal social influence task

In the Reciprocal social influence task, the agent (computer or robot) made final decisions in the observation turns. In this regard, we implemented two distinct types of behavior in two different experimental conditions (Susceptible and Unsusceptible). These two conditions were implemented in two consecutive experimental blocks, interleaved by a Transition block (see the next paragraph, "transition and final blocks"). The distributions of the final responses in the two different conditions were systematically manipulated in terms of *influence* (i.e., shift towards the participant's estimate in the final decision) of the agent.

This could range from 0 to 1, where an influence of 1 refers to a final decision coinciding with the participant's perceptual estimate and an influence of 0 corresponds to a final decision coinciding with the agent's own estimate.

On average, the agents' influence in participants was 0.64 (within-subject SD = 0.26, between-subject SD = 0.04) and 0.18 (within-subject SD = 0.16, between-subject SD = 0.02) in Susceptible and Unsusceptible conditions, respectively. More specifically, in the Susceptible condition trial-by-trial influence was chosen with a probability of 0.45 from a uniform distribution in the interval [0.75, 1], with a probability of 0.35 randomly from a uniform distribution in the interval [0.5, 0.75], with a probability of 0.15 randomly from a uniform distribution in the interval [0.25, 0.5] and with a probability of 0.05 randomly from a uniform distribution in the interval [0, 0.25]. In the Unsusceptible condition the agent's level of *influence* in each trial was chosen with a probability of 0.80 from a uniform distribution in the interval [0, 0.25], with a probability of 0.15 randomly from a uniform distribution in the interval [0.25, 0.5] and with a probability of 0.05 randomly from a uniform distribution in the interval [0.5, 0.75].

These intervals and distributions were chosen to balance variability and credibility of the agents' final decisions. On the one hand, we wanted participants to believe that the agent was dynamically adjusting its level of influence based on informational and normative considerations and that their final decisions were not pre-determined for experimental purposes. On the other hand, we wanted to ensure that participants could detect the change of behavior of the agent across Susceptible and Unsusceptible conditions. These algorithms were tested in a pilot study, where participants, in the final debriefing, confirmed that they spotted the partner's behavioral switch and that they did not perceive the partner's behavior as implausible or pre-programmed.

Between the two main experimental blocks (Susceptible and Unsusceptible), we introduced a smooth transition in the behavior of the agent (computer or robot). Concerning the transition from the Susceptible to the Unsusceptible block, the other agent's influence in the 11 observation turns was chosen twice from a uniform distribution in the interval [0.75, 1], then twice from a uniform distribution in the interval [0.5, 0.75], then three times from a uniform distribution in the interval [0.25, 0.5] and eventually four times from a uniform distribution in the interval [0, 0.25], in this exact order. Conversely, in the transition from the Unsusceptible to the Susceptible block, the trial-by-trial magnitude of the partner's influence was chosen from the same distributions, but it was presented in the opposite order, moving sequentially from the interval [0, 0.25] to the interval [0.75, 1].

The Final block included 11 observation turns in which the partner's trial-by-trial influence was picked from the same probabilistic distribution of the last condition faced in the Reciprocal social influence task.

## QUANTIFICATION AND STATISTICAL ANALYSIS

Most of the analyses reported in this work focus on examining the variation of dependent variables related to social influence (i.e., influence, estimation error) as a function of exogenous experimental factors (i.e., experimental groups, experimental conditions) and endogenous predictors (i.e., distance from other agent's response, performance ratings). First, we used mixed-effect models on trial-by-trial data with random effects at the subject level. In all the models of the paper, random effects have been applied to the intercept to adjust for the baseline level of influence of each subject and model intra-subject correlation of repeated measurements. Specification and results of each model have been described in detail in the supplemental information. Moreover, throughout the paper we also directly compared individual variables (e.g., influence, estimation error, performance ratings) across experimental conditions and groups. Since these individual variables occasionally show some degree of skewness and, in some conditions, show a violation of the normality distribution assumption, we used non-parametric tests (Wilcoxon signed-rank test; Wilcoxon rank-sum test) through the entire paper for consistency. All tests are two-tailed and report z statistic, p. value and effect sizes (r, $\eta^2$). For the same reason, we used non-parametric correlation tests (Spearman's rank correlation). The formulas used for the calculation of the effect sizes can be found in Cohen (2008) and Fritz et al. (2012): $r = Z/\sqrt{N}$ (total number of observations); $\eta^2 = Z^2/N$. Figures use asterisks to express the significance of the relevant statistical tests (***p < 0.001, **p < 0.01, *p < 0.05, ns: not significant). Error bars in all the figures represent between-subject standard error of the mean. All analyses include the entire sample of 50 subjects and all trials of the three experimental tasks.